\documentclass{article}
\usepackage{ijcai17}
\usepackage{times}
\usepackage{graphicx}
\usepackage{float}
\usepackage{subfig}
\usepackage{amsmath,amssymb} 
\usepackage{hyperref}
\usepackage{float}
\usepackage[table]{xcolor}
\usepackage{setspace}
\usepackage[font=small,labelfont=bf]{caption}
\definecolor{Gray}{gray}{0.85}

\newcommand\Mark[1]{\textsuperscript#1}
\restylefloat{table}

\begin{document}

\title{MAT: A Multimodal Attentive Translator for Image Captioning}

\author{
	Chang Liu\Mark{1}\Mark{,}\Mark{4}, Fuchun Sun\Mark{1}, Changhu Wang\Mark{2}, Feng Wang\Mark{3}\Mark{,}\Mark{4}, Alan Yuille\Mark{4}\\
	\Mark{1}Department of Computer Science, Tsinghua University \\
	\Mark{2}Toutiao AI Lab,
	\Mark{3}Department of Electronic Engineering, UESTC\\
	\Mark{4}Cognitive Science \& Computer Science, Johns Hopkins University\\
	\Mark{1}\{cliu13@mails, fcsun@mail\}.tsinghua.edu.cn,\\
	\Mark{2}wangchanghu@toutiao.com,
	\Mark{3}feng.wff@gmail.com,
	\Mark{4}alan.yuille@jhu.edu
}
\maketitle

\begin{abstract}
In this work we formulate the problem of image captioning as a multimodal translation task. Analogous to machine translation, we present a sequence-to-sequence recurrent neural networks (RNN) model for image caption generation. Different from most existing work where the whole image is represented by convolutional neural network (CNN) feature, we propose to represent the input image as a sequence of detected objects which feeds as the source sequence of the RNN model. In this way, the sequential representation of an image can be naturally translated to a sequence of words, as the target sequence of the RNN model. To represent the image in a sequential way, we extract the objects features in the image and arrange them in a order using convolutional neural networks. To further leverage the visual information from the encoded objects, a sequential attention layer is introduced to selectively attend to the objects that are related to generate corresponding words in the sentences. Extensive experiments are conducted to validate the proposed approach on popular benchmark dataset, i.e., MS COCO, and the proposed model surpasses the state-of-the-art methods in all metrics following the dataset splits of previous work. The proposed approach is also evaluated by the evaluation server of MS COCO captioning challenge, and achieves very competitive results, e.g., a CIDEr of 1.029 (c5) and 1.064 (c40).
\end{abstract}

\section{Introduction}
Image captioning is a challenging problem. Unlike other computer vision tasks such as image classification and object detection, image captioning requires not only understanding the image, but also knowledge of natural language. Early methods for image captioning either explored template-based, e.g., ~\cite{mitchell2012midge,elliott2013image} or retrieval-based approaches~\cite{gong2014improving,kuznetsova2014treetalk}. However, the language models were usually heavily hand-designed, and found it hard to generate novel sentences with new compositions.

Inspired by the success of sequence-to-sequence machine translation~\cite{sutskever2014sequence}, based on recurrent neural networks, recent approaches for image captioning brought new insights by using a two-stage `encoding' and `decoding' technique~\cite{Donahue_2015_CVPR,Karpathy_2015_CVPR,Vinyals_2015_CVPR}. The common idea of these approaches is to use the whole CNN feature of the image as the `source' input, to replace the words of the `source language' in the translation task. The caption is then generated by conditioning the output words on the CNN feature of the image. 
One problem these approaches suffer from is the imbalance of the encoded visual information (representation of the image) and the language part (representation of words), because in an RNN the image in the `source language' only provides the CNN feature of image at one time step only, i.e., the initial step, but the words in the `target language' contribute at multiple time steps. Since the power of RNN lies in its capability to model the contextual information between time steps~\cite{hochreiter1997long}, such image representation weakens the RNN's memory of the visual information as it contains no temporal concept.

To encode more visual information, some approaches, e.g., \cite{mitchell2012midge,yang2011corpus} represent the input image in terms of the objects it contains. One strategy that some early work leverages is to construct inputs for the language model using object-attribute pairs, with objects and attributes represented by descriptive words, e.g., \cite{mitchell2012midge,yang2011corpus}, in which, however, the visual features of objects are actually not utilized. Another strategy some recent methods such as \cite{Fang_2015_CVPR,wu2016value,you2016image} proposed is to leverage visual descriptors of image parts to enhance the visual information using weakly supervised approach. More recently, some approaches leverage the attention mechanism to encode more visual information \cite{xu2015show,you2016image}, in which the models learn to fix their gaze on salient regions when generating the corresponding words.

In this work, we follow the basic strategy of CNN and RNN captioning model, but expand the image representation from one single CNN feature at a single time step in RNN, to a sequence of objects at multiple time steps. In this way, analogous to machine translation, image captioning is formulated as translating a visual language (a sequence of objects) to a natural language (a sequence of words). This idea has several advantages. Firstly, we can leverage object detection techniques to encode more visual information in the visual language; secondly, the sequential representation of the image accords more with the temporal concept of RNN models, and makes the two sides of translation more balanced. On the basis of the multi-modal sequence translation model, we also introduce a `sequential attention' mechanism, where our model can learn to  distribute its attention to different objects in the image. It should be noted that the attention mechanism in our model differs from previous ones because the attention in our model works on the basis of sequential representation of the image, by taking all encoding objects information into account when generating words; while the attention in previous work focuses on the feature map of the whole image representation, and sequential representation of the image is not explored.

To summarize, the main contributions of this work are:

$\bullet$We present a novel multimodal translation model for image captioning, which translates the `visual language' of a sequence of objects to a natural language with a sequence of words. This sequential formulation ensures the balance between the encoding information and the decoding information of the RNN.

$\bullet$We introduce a `sequential attention' layer, which learns to consider all the encoder information at each time steps during decoding with different weights.

$\bullet$We quantitatively validate our model on benchmark dataset, and surpass the state-of-the-art methods in all metrics. For example, on MS COCO Captioning Challenge evaluation server, our model achieves a CIDEr of 1.029 (c5) and 1.064(c40), while the second best achieves 0.965 (c5) and 0.969 (c40). In addition, we also evaluate our method using the more recent metric SPICE\cite{anderson2016spice}, which accords more with human judgments, and our model achieves 18.9 (c5).

\section{Related Work}
\noindent\textbf{CNN+RNN based captioning.} A typical captioning way is to combine CNN and RNN, where CNN is used to extract the feature of the whole image, and RNN to construct the language model. For example, Vinyals \emph{et al}.~\cite{Vinyals_2015_CVPR} proposed an end-to-end model composed of a CNN and an RNN. The model is trained to maximize the likelihood of the target sentence given the CNN feature of the training image at the initial time step. Mao \emph{et al}.~\cite{Mao_2015_ICCV} presented an m-RNN model, where the CNN feature of the image is fed into the multimodal layer after the recurrent layer rather than the initial time step. Similar work that utilizes CNN and RNN to generate descriptions includes~\cite{Chen_2015_CVPR,Karpathy_2015_CVPR}. However, most of above methods represent the image in a static form, such as a 4096-d CNN feature vector. Although this feature can well represent an image, it is insufficient for the sequential RNN model. That is because such a feature only provides the encoding phase of the RNN model with a single time-step data, leaving the rest of the model to the decoding phase where words in the caption are used.

\noindent\textbf{Sequence based captioning.} Several algorithms extend the image representation from a single time-step data to multiple time-step data in image captioning~\cite{Donahue_2015_CVPR} and video description~\cite{Donahue_2015_CVPR,venugopalan2015sequence}. In image captioning, \cite{Donahue_2015_CVPR} proposed to feed the image feature to the RNN at each time step; and in video description, Venugopalan \emph{et al}. first computed the CNN feature for each video frame, and then fed the mean pooling of these features to the RNN at every time step in the encoding phase. Note that in these algorithms, the inputs to RNN during encoding are the same at each time step; they are the duplicate features of the image~\cite{Donahue_2015_CVPR}. Feeding the same information at each time step cannot make the networks to obtain more contextual information during encoding, but only strengthens the same concept over and over. On the other hand, instead of extending the representation with the same inputs, \cite{Donahue_2015_CVPR} and \cite{venugopalan2015sequence} proposed to extract the CNN feature for each video frame, and feed them to the RNN model one frame per time step, to avoid from using repeated feature inputs. Our work is different from above two approaches, for we focus on image captioning with object-level representation, rather than a video description with frame-level representation.

\noindent\textbf{Attention based captioning.} Visual attention has been proved as an effective way for the task of image captioning \cite{xu2015show,yang2016review,you2016image,liu2016attention}. These attention based captioning models are capable of learning where to attend in the image when generating the target words.They either learn the distribution of spatial attention from the last convolutional layer of the convolutional neural network\cite{xu2015show,yang2016review}, or learn the distribution of semantic attention from the visual attributes that are learnt from weakly annotated images from social media\cite{you2016image}. While these methods show the effectiveness of the attention mechanism, the contextual information in the encoding sequence is not explored. Our attention layer differs as it takes a sequential form, where each hidden state of the encoding stage contributes to generating the decoding words.

\begin{figure*}[t]
	\begin{center}
		\includegraphics[width=0.7\linewidth]{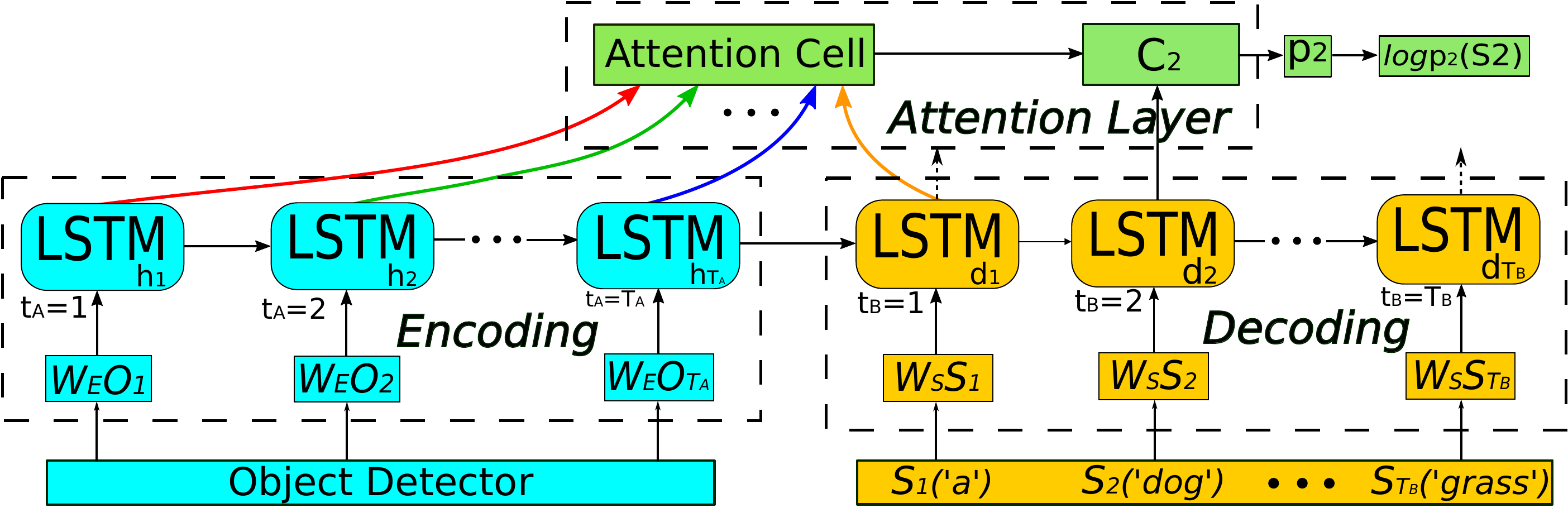}
	\end{center}
	\caption{Overview of the Model. The source sequence is represented with the embedding of object in a hidden space; the target sequence is represented with the embedding of the words in the same space. The mapping from the source to the target is modeled with two LSTM units in an unrolled version. An attention layer computes the context vector  $C_{t}$ from all the encoding hidden states and decoding hidden state from last timestep The target word is generated by a softmax over the vocabulary given the attention context vector. Better view in color.}
	\label{fig:overview}
\end{figure*}

\section{Proposed Method}
In the following sections, we first formulate the problem as a multimodal translation problem\ref{subsec:fomulation}. We then present model in detail in Section~\ref{subsec:model}. After which we elaborate the attention layer in Section \ref{subsec:attention_layer}. And Section \ref{subsec:training_and_inference} states  the training and inferences details of the model.
\subsection{Formulation}
\label{subsec:fomulation}
In previous work with CNN+RNN solutions, the core idea is usually to maximize the probability of the description given the input image:
\begin{equation}
\log{p(S|I)} = \sum_{t=1}^{N}\log{p(S_t|I,S_{1:t-1})}
\end{equation}
where $I$ represents the image, $S_i$ is the $i$th word in sentence $S$, and $p(S_t|I,S_{1:t-1})$ is the probability of generating word $S_t$ given the image and previous words $S_{1:t-1}$. A common representation of the image is a CNN feature vector, and the recursive language part is usually modeled with recurrent neural networks, where an RNN unit considers the following two data as inputs: (1) input at current time step $t$, and (2) output from the previous time step $t-1$.

In this work, we formulate image captioning as a multimodal translation task, i.e., translating a visual language (a sequence of objects) to a natural language (a sequence of words). The core idea is to use RNN to model the translation process, by feeding one object at a time to an RNN unit during encoding, and one word at a time during decoding. Besides this multimodal translation scheme, we also introduce an attention context vector $C_t$ to generate the word during decoding, which comes from the proposed sequential attention layer. The attention layer shares similar spirit from the literature of machine translation\cite{vinyals2015grammar,luong-pham-manning:2015:EMNLP,bahdanau2014neural}, where the core motivation is to consider all the hidden states of the encoding when computing the attention vector. Thus, we call our model Multimodal Attentive Translator, MAT for short.

To be specific, given an image $I$, we use $seq(I)$ to denote its sequential representation, which contains a sequence of representations $seq(I)=\{O_1,O_2,...,O_{T_A}\}$, where $O_1$ to $O_{T_A-1}$ are the object representations and the last item $O_{T_A}$ encodes the global environmental information by feeding CNN feature of the whole image.Then the RNN takes in $seq(I)$ by encoding each object into a fixed length vector, and is recursively activated at each time step. We denote the attention vector as $C_{t_b}$ at decoding time step $t_b$, where $t_b$ ranges from 1 to $T_B$, i.e., the decoding length. The sentence is generated by conditioning the outputs given attention context vector $C_{t_B}$  (Eqn.~\ref{eq:generation}), where $C_{t_B}$ is computed from the attention layer (Eqn.~\ref{eq:attention}).
\vspace{-0.1in}
\begin{equation}
\label{eq:generation}
\log{p(S|seq(I))}=\sum_{t_B=1}^{T_B}{\log{p(S_{t_B}|C_{t_B},S_1,...,S_{t_B-1})}}
\end{equation}
\vspace{-0.1in}
\begin{equation}
\label{eq:attention}
C_{t_B} = ATT(H, d_{t_B-1}),
\end{equation}

where $H$ denotes all the encoding hidden states $h_1, h_2, ..., h_{T_A}$, and $d_{t_B-1}$ denotes the decoding hidden state of last time step. And we have encoding states and decoding states computed from:

\begin{equation}
\label{eq:encoder}
h_{t_A} = RNN_{en}(seq(I)_{t_A}, h_{t_A-1}), t_A = 1,2, ..., T_A
\end{equation}

\begin{equation}
\label{eq:decoder}
d_{t_B} = RNN_{de}(S_{t_B}, d_{t_B-1}), t_B = 1,2, ..., T_B
\end{equation}

\subsection{Seq2Seq Multimodal Translator}
\label{subsec:model}
\noindent\textbf{Source sequence representation.}
The source sequence is represented by the objects from the image. We first use object detectors to locate the objects and extract $D_o$-dimension CNN features, denoted as CNN($O_{t_A}$) for the $t_A$-th object $O_{t_A}$. Then object features are mapped into an $H$-dimension hidden space with embedding matrix $W_E$, with $\mathbb{R}^{H\times{D_o}}$ dimension. The source sequence of the RNN is represented as:
\begin{equation}
x_{t_A}={W_E}{\text{CNN}(O_{t_A})}, \quad t_A\in\{1,2,...,T_A\},
\label{eq:source_sequence}
\end{equation}
where $t_A$ is the encoding timestep of the network, and $T_A$ is the total length of encoding state.

\begin{figure}
\begin{center}
		\includegraphics[width=0.5\linewidth]{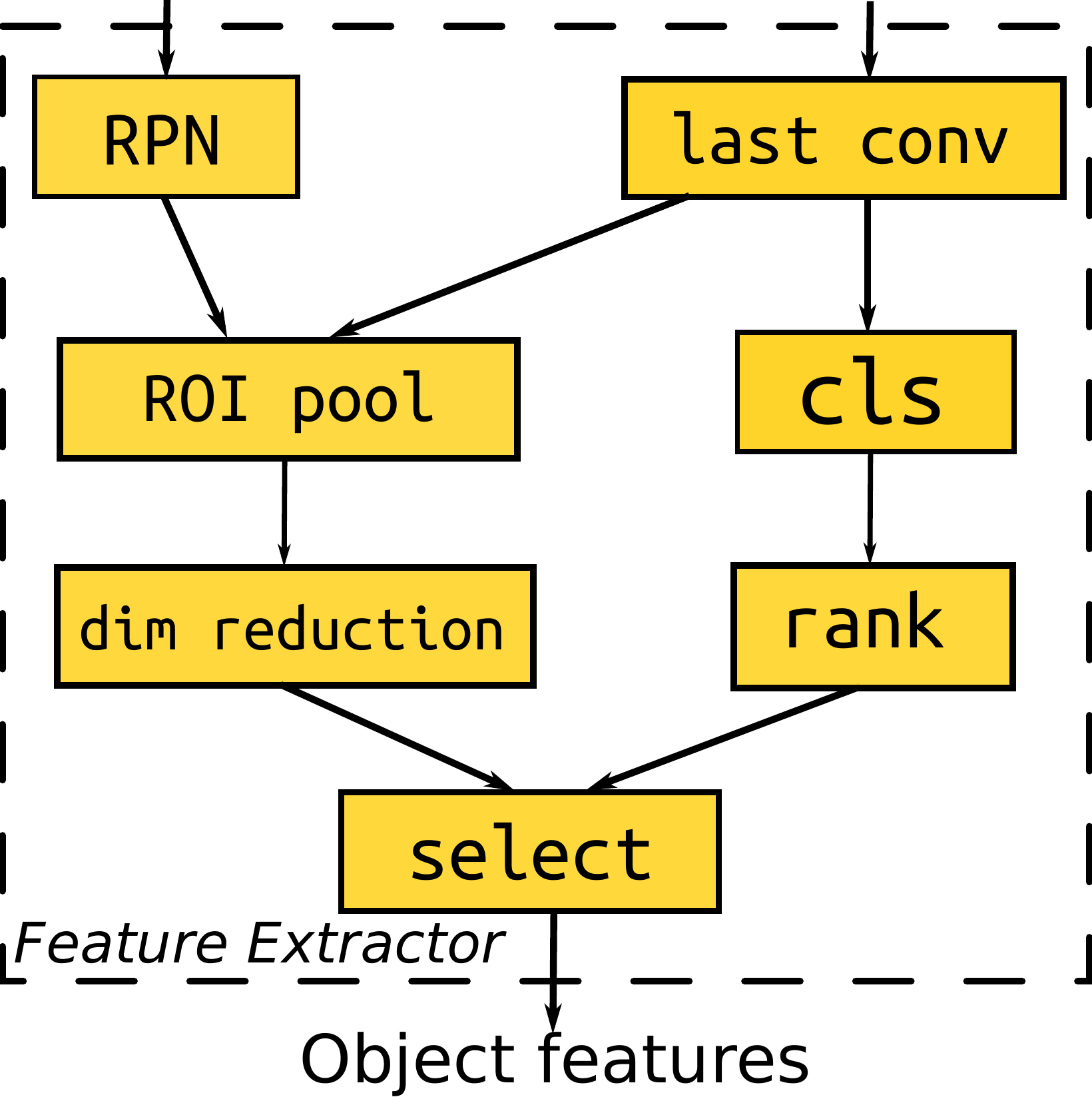}
	\end{center}
	\caption{Object feature extractor of the model. We extract object features from the last convolutional layer of detection network using roi pooling. The pooled features are selected by the detection score (i.e., cls) from highest to lowest.}
	\label{fig:feature_extractor}
\end{figure}

In general, any object detection method can be leveraged to generate the previous mentioned object features. The feature extractor is shown in Figure\ref{fig:feature_extractor}, and it can be inserted to any RPN based detection network. In practical, we adopt the R-FCN\cite{li2016r} detection network, where we pool the object features from the last convolutional layer using \textit{roi pooling}. Details refer to \ref{subsec:training_and_inference}.

\noindent\textbf{Target sequence representation.}
The target sequence is represented with a set of words ${S_1,...,S_N}$ in the sentence $S$. Each word $S_{t_B}$ is represented as a `one-hot' vector, with a dimension $\mathbb{R}^{D_s}$ equal to the vocabulary size. We add special START word and END word to denote the start and the end of the sentence. The start word and end word are also included into the vocabulary of the model. Then the words are mapped to the same hidden space with word embedding matrix $W_S$, with a dimension of $\mathbb{R}^{H\times{D_s}}$. Thus the target sequence is represented by:
\begin{equation}
x_{t_B}={W_S}{S_{t_B}}, \quad t_B\in\{{1,2,...,T_B}\},
\label{eq:target_sequence}
\end{equation}
where $t_B$ is the decoding timestep of the network, and $T_B$ is the total length of decoding stage.

\noindent\textbf{RNN translation from source to target.}
To model the translation from the source sequence to the target sequence, we leverage the long-short term memory (LSTM), a specific unit of RNN, to avoid the gradient exploding and vanishing problem of the network~\cite{hochreiter1997long,hochreiter2001gradient}.

LSTM takes in the output of the previous time step, as well as the input at the current time step, as the inputs of the current unit. To better illustrate the idea of this recursive process, we unroll the encoding $LSTM_1$ and decoding $LSTM_2$ along the time dimension, by copying the LSTM unit at each time step, as shown in Fig.~\ref{fig:overview}. The core of an LSTM unit is a memory cell $c$, which is controlled by several gates. The activation of each gate determines whether the corresponding input is accepted or rejected. Thus the mapping from the source (Eqn.~\ref{eq:source_sequence}) to the target (Eqn.~\ref{eq:target_sequence}) is formulated by the following equations:
\vspace{-0.1in}
\begin{equation}
	i_t=\sigma(W_{xi}x_t+W_{hi}h_{t-1}+b_i), 
	\label{eq:i}
\end{equation}
\begin{equation}
	f_t=\sigma(W_{xf}x_t+W_{hf}h_{t-1}+b_f),
	\label{eq:f}
\end{equation}
\begin{equation}
	o_t=\sigma(W_{xo}x_t+W_{ho}h_{t-1}+b_o),
	\label{eq:o}
\end{equation}
\begin{equation}
	g_t=\sigma(W_{xg}x_t+W_{hg}h_{t-1}+b_g),
	\label{eq:g}
\end{equation}
\begin{equation}
	c_t=f_t\odot{c_{t-1}}+i_t\odot{g_t},
	\label{eq:c}
\end{equation}
\begin{equation}
	h_t=o_t\odot{\phi(c_t)},
	\label{eq:h}
\end{equation}

where $t$ ranges from the start of the source sequence to the end of the target sequence in all equations; $i_t$, $f_t$, $o_t$, and $g_t$ represent the input gate, forget gate, output gate, and input modulation gate at time step $t$, respectively; $c_t$ and $h_t$ are the memory cell and the hidden state; $W_{ij}$ represents the connection matrix and $b_j$ is the bias; $\sigma$ is the sigmoid non-linearity operator, and $\phi$ is the hyperbolic tangent non-linearity; $\odot$ is the element-wise multiplication operator. 

\subsection{Sequential Attention Layer}
\label{subsec:attention_layer}
Although the decoding hidden state of LSTM unit comes from previous encoding hidden state, the states that are far from the current decoding state may contribute little to generate the word. In other words, objects that appear at the very beginning of the source sequence may be well related to the word at the decoding step. For example, a dog detected with highest score will appear at the first time step of the encoding procedure, but the word "dog" may be at the end of the decoding time step, if the sentence is "A man is playing with a dog." Driven by the idea, we design a attention context vector, which lets the model generate words considering all hidden encoding states, computed from the proposed attention layer. The attention layer thus takes inputs from hidden states $H$, as well as previous decoding state $d_{t_B}$, and outputs the attention context vector, as shown in Figure. \ref{fig:attention}. Specifically, the distribution $p_{t_B+1}$ is computed from the context vector $C_{t+1}$, which derives as (for all the equations we omit bias term for simplicity):
\begin{figure}[t]
	\begin{center}
		\includegraphics[width=1.0\linewidth]{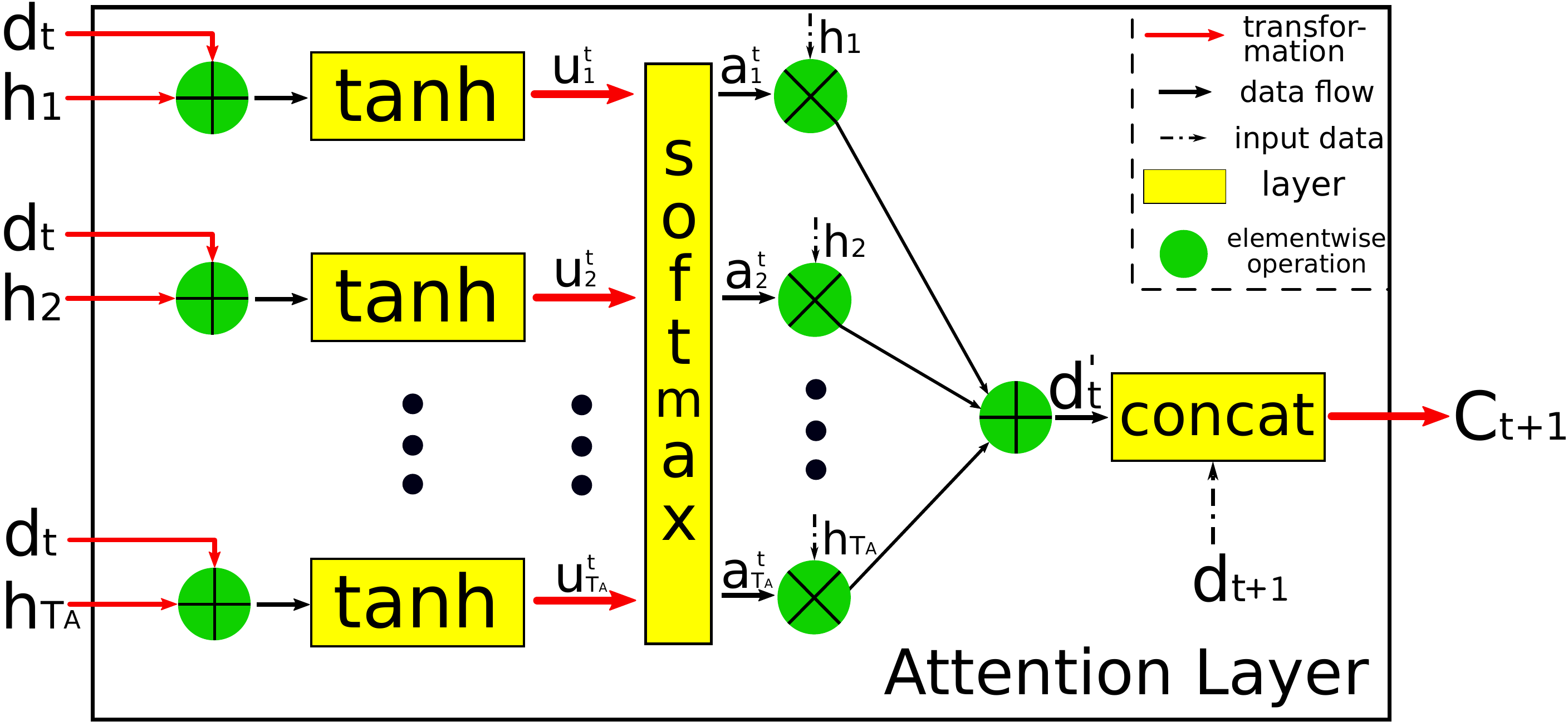}
	\end{center}
	\caption{Attention Layer. At decoding timestep $t=t_B$, the attention layer computes the attention context vector $C_{t+1}$ from all the encoding hidden states $h_1, ..., h_{T_A}$, as well as previous decoding state $d_t$. Better view in color.}
	\label{fig:attention}
\end{figure}
\begin{table*}[!htbp]
	\fontsize{9}{9}\selectfont
	\caption{Comparison results on MSCOCO on 5000 testing images following previous work. The highest score is labeled as bold. Our model is noted as `MAT(ours)' in gray. All metrics are reported using c5 references. Note SPICE metric is recently proposed, and lack refereed methods to compare on.}
	\centering
	\begin{tabular}{|*{9}{c|}}	\hline
		\multicolumn{1}{|c|}{} & \multicolumn{7}{c}{MS COCO, 5000 testing images} & \multicolumn{1}{c|}{} \\\hline
		Methods & CIDEr & METEOR & RougeL & BLEU-1 & BLEU-2 & BLEU-3 & BLEU-4 & SPICE\\\hline
		Berkeley LRCN~\cite{Donahue_2015_CVPR} & - & - & - &  0.628 & 0.442 & 0.304 & 0.210 & - \\\hline
		DeepVS~\cite{Karpathy_2015_CVPR} & 0.660 & 0.195 & - & 0.625 & 0.450 & 0.321 & 0.230 & - \\\hline		
		Attention~\cite{xu2015show} & - & 0.230 & - & 0.718 & 0.504 & 0.357 & 0.250 & -\\\hline
		Reviewnet~\cite{yang2016review} & 0.886 & 0.237 & -&  - & - & - & 0.290 & - \\\hline
		Semantic~\cite{you2016image} & - & 0.243 & - &  0.709 & 0.537 & 0.402 & 0.304 & - \\\hline
		\rowcolor{Gray}MAT(ours) & \textbf{1.058} & \textbf{0.258} & \textbf{0.541} & \textbf{0.731} & \textbf{0.567} & \textbf{0.429} & \textbf{0.323} & \textbf{18.9} \\\hline
	\end{tabular}
	\label{tabel:test_5000}
\end{table*}

\begin{table*}[!htbp]
	\caption{Comparison results on MS COCO Caption Challenge 2015, using MS COCO evaluation server. All metrics are reported using c5 references.}
	\centering
	\begin{tabular}{|*{8}{c|}}	\hline
		\multicolumn{1}{|c|}{} & \multicolumn{6}{c}{MS COCO Captioning Challenge, 40775 images (c5)} & \multicolumn{1}{c|}{} \\\hline
		Methods & CIDEr & METEOR & RougeL & B-1 & BLEU-2 & BLEU-3 & BLEU-4 \\\hline
		Google NIC~\cite{Vinyals_2015_CVPR} & 0.943  & 0.254 & 0.530 & 0.713 & 0.542 & 0.407 & 0.309  \\\hline
		MSR Captivator~\cite{DBLP:conf/acl/DevlinCFGDHZM15} & 0.931 & 0.248 & 0.526 & 0.715 & 0.543 & 0.407 & 0.308\\\hline
		Berkeley LRCN~\cite{Donahue_2015_CVPR} & 0.921 & 0.247 & 0.528 & 0.718 & 0.548 & 0.409 & 0.306    \\\hline
		m-RNN~\cite{Mao_2015_ICCV} & 0.917 & 0.242 & 0.521 & 0.716 & 0.545 & 0.404 & 0.299    \\\hline
		MSR~\cite{Fang_2015_CVPR}& 0.912& 0.247& 0.519 & 0.695 & 0.526 & 0.391 & 0.291    \\\hline
		ACVT~\cite{wu2016value}& 0.911 & 0.246 & 0.528 & 0.725 & 0.556 & 0.414 & 0.306    \\\hline
		Attention~\cite{xu2015show}& 0.865& 0.241 & 0.516 & 0.705 & 0.528 & 0.383 & 0.277    \\\hline
		Reviewnet~\cite{yang2016review}& 0.965 & 0.256 & 0.533 & 0.720 & 0.550 & 0.414 & 0.313   \\\hline
		Semantic~\cite{you2016image}& 0.943 & 0.250 & 0.530 & 0.713 & 0.542 & 0.407 & 0.309   \\\hline
		\rowcolor{Gray}MAT(ours)& \textbf{1.029} & \textbf{0.258} & \textbf{0.540} & \textbf{0.734} & \textbf{0.568} & \textbf{0.427} & \textbf{0.320}  \\\hline
	\end{tabular}
	\label{table:test_2014}
\end{table*}
\vspace{-0.1in}
\begin{equation}
	p_{t_B+1} = Softmax(W_C(out_{t_B+1})
\end{equation}
\begin{equation}
	out_{t_B+1} = Concat(d_{t_B+1}, {d_{t_B}{'}})
\end{equation}
\begin{equation}
	{d_{t_B}{'}}= \sum_{i=1}^{T_A}a_{i}^{t_B}h_i
\end{equation}
\begin{equation}
	a_{i}^{t_B} = Softmax(u_{t_B})_i
\end{equation}
\begin{equation}
	u_{i}^{t_B} = V^Ttanh(W_Hh_i+W_Dd_{t_B})
\end{equation}
where $V$,  $W_H$, $W_D$ and $W_C$ are the parameters of the layer. $V \in{\mathbb{R}^{H\times{1}}}$ dimension vector; $W_H, W_D \in{\mathbb{R}^{H\times{H}}}$; $W_C \in{\mathbb{R}^{2H\times{H}}}$. The vector $u_{t_B}$ is of length $T_A$, which represents how much attention the model pays to each encoding hidden state $h_{t_A}$ when generating the attention context vector. $u_{t_B}$ is then used to compute a new state vector, ${d_{t_B}{'}}$. We then concatenate ${d_{t_B}}^{'}$ and current decoding hidden state $d_{t_{B+1}}$, with which we feed as the hidden state to make prediction over the vocabulary using softmax to generate the word.

\subsection{Training and Inference}
\label{subsec:training_and_inference}
\noindent\textbf{Training.} To obtain the sequential representation of the image, we run object detectors on the image. Specifically, we leverage R-FCN\cite{li2016r} trained on MS COCO dataset, using the Resnet101 convolutional architecture\cite{he2016deep}. We use a threshold of 0.5 to detect the objects in the image, and the objects are ordered according to their detection scores. We use two different LSTMs for encoding and decoding. The hidden state size is set to 512. To cope with variable length of both the source sequence and the target sequence for batch training, we leverage a bucket and padding method, where the sequences are split to different buckets and zero padded to bucket length according to the length of the source sequence as well as the length of the target sequence. Specifically, in training, we use four buckets, i.e., \{(2, 10), (4,15), (6,20), (8,30)\}. For example, given a training sample with 5 objects and 10 words, it belongs to the third bucket (6,20) since although 10 words $<$ 15 threshold, the object number 5 exceeds threshold 4, thus the second bucket doesn't fit. We then zero padding the words from 10 to 15, and the objects from 5 to 6. Our loss is the sum of the negative log likelihood of the generated words at decoding time steps, i.e.: 
\begin{equation}
	Loss = -\sum_{t_B=1}^{T_B}\log p_{t_B}(S_{t_B})
\end{equation}
We use SGD with batch size of 64 to train the network. The learning rate is set to 0.1, and halved when training loss stops to decrease. To avoid overfitting, we leverage dropout at 0.5 for all layers, and early stops the training on validation split with 5000 images. On a Titan X Maxwell computer, the training process takes about 12 hours.

\noindent\textbf{Inference.} We first run object detectors on the testing image, and the feature representations of the objects are fed in to the encoding parts of the model. We use BeamSearch of size 20, which considers iteratively the best $b$ candidates to generate next word. The sentence generation stops when it generates the special END word.

\section{Experimental Results}
\label{section:experiments}
\subsection{Datasets and Evaluation Measurements}
\label{subsection:datasets}
\noindent\textbf{MSCOCO}~\cite{lin2014microsoft} contains 82,783 training, 40,504 validation and 40,775 testing images, which are withheld in MS COCO server. To compare with previous methods, we follow the split from previous work~\cite{Karpathy_2015_CVPR,xu2015show}, i.e., we use 5000 images for validation and 5000 images for testing from the 40504 validation set. Moreover, we also compare with the state-of-the-art methods listed in the leader board on the MSCOCO  website, where 40,755 images are withheld for testing (ground truth sentences unavailable). We follow previous work to generate the vocabulary of the model, that is, we first count all the occurrence number of all the words and filter the words which occurrence number is less than 5. The final vocabulary size is 8791  .
\begin{table*}[ht]
	\caption{Comparison between MAT and baseline methods. Baseline 1 is CNN+RNN method. Baseline 2 is Seq2Seq method, i.e., using MAT model but without attention layer.}
	\centering
	\begin{tabular}{|*{9}{c|}}	\hline
		\multicolumn{1}{|c|}{} & \multicolumn{7}{c}{MS COCO, 5000 testing images} & \multicolumn{1}{c|}{} \\\hline
		Methods & CIDEr & METEOR & RougeL & BLEU-1 & BLEU-2 & BLEU-3 & BLEU-4 & SPICE\\\hline
	    Baseline 1& 0.927 & 0.230 & 0.516 & 0.705 & 0.504 & 0.357 & 0.250 & 17.2\\\hline
		Baseline 2& 0.961 & 0.245 & 0.519 & 0.711 & 0.542 & 0.403 & 0.298 & 17.8 \\\hline
		\rowcolor{Gray}MAT(ours) & \textbf{1.058} & \textbf{0.258} & \textbf{0.541} & \textbf{0.731} & \textbf{0.567} & \textbf{0.429} & \textbf{0.323} & \textbf{18.9} \\\hline
	\end{tabular}
	\label{tabel:baseline}
\end{table*}

\noindent\textbf{Evaluation measurements}: We use the public available MS COCO evaluation toolkit\footnote{https://github.com/tylin/coco-caption.git} to evaluate our model, which computes BLEU~\cite{papineni2002bleu}, CIDEr~\cite{vedantam2014cider}, METEOR\cite{banerjee2005meteor}, and RougeL\cite{lin2004rouge}. In addition, we evaluate our model with the recent metric SPICE\cite{anderson2016spice}, which accords more with human judgments. For all the metrics, the higher the better.

\subsection{Overall Comparison with the State of the Arts}
\label{sec:overall_comparison}

\noindent\textbf{MSCOCO}: The results are shown in Table~\ref{tabel:test_5000} and Table~\ref{table:test_2014}, which show the performance of our model compared to state-of-the-art published methods. From Table~\ref{tabel:test_5000} and Table~\ref{table:test_2014} it can be seen that our model outperforms state-of-the-art in all evaluation measurements, which implies the validity and effectiveness of proposed model.

In consideration of the latent possibility that the metrics used by MS COCO evaluation server may drift away from human judgments, although hiring human to evaluate and compare the generated sentences is a more reliable and solid way, it's very money and time consuming and thus is not always practical. Nonetheless, in addition to validate on MS COCO evaluation server, we also evaluate our model using the recent proposed metric SPICE~\cite{anderson2016spice}, which is more consistent with human judgments and gives a better idea of the performance of language generating models. We achieve a SPICE of 18.9(c5), using public available SPICE evaluation tool\footnote{https://github.com/peteanderson80/coco-caption.git}. Given no contemporary work reports SPICE on MS COCO 5000 testing split, and performances reported in the SPICE paper~\cite{anderson2016spice} use C40 testing set obtained from MS COCO organizers, which is inaccessible to us and thus makes our C5 performance incomparable, we decide, nonetheless, to publish our C5 SPICE performance on 5000 testing split, in order to provide a comparable number for future work.

\subsection{Comparison with Baseline Methods}
We compare our MAT model with two baselines: 1) vanilla CNN+RNN captioning model, where single CNN image feature is used; 2) Seq2Seq model without attention layer, i.e. take out attention layer from MAT model. The experiment settings, i.e., training and inference details for baseline 1 and baseline 2 are consistent with that of MAT model. The results are shown in Table\ref{tabel:baseline}. Two conclusions can be drawn from the Table. Firstly, from the comparison between baseline 1 and baseline 2, it shows that the sequential representation of the image indeed improves the performance compared to the single CNN feature of the representation, which accords with idea of encoding more visual information of the source sequence of the RNN. Secondly, when comparing MAT results to both baseline1 and baseline2, it can be drawn that the proposed sequential attention layer further improves the performance by a large margin. These two facts indicate the effectiveness of our multimodal translator with sequential attention layer.

\subsection{Qualitatively Evaluation}
\begin{figure}[ht]
	\begin{center}
		\includegraphics[width=1.0\linewidth]{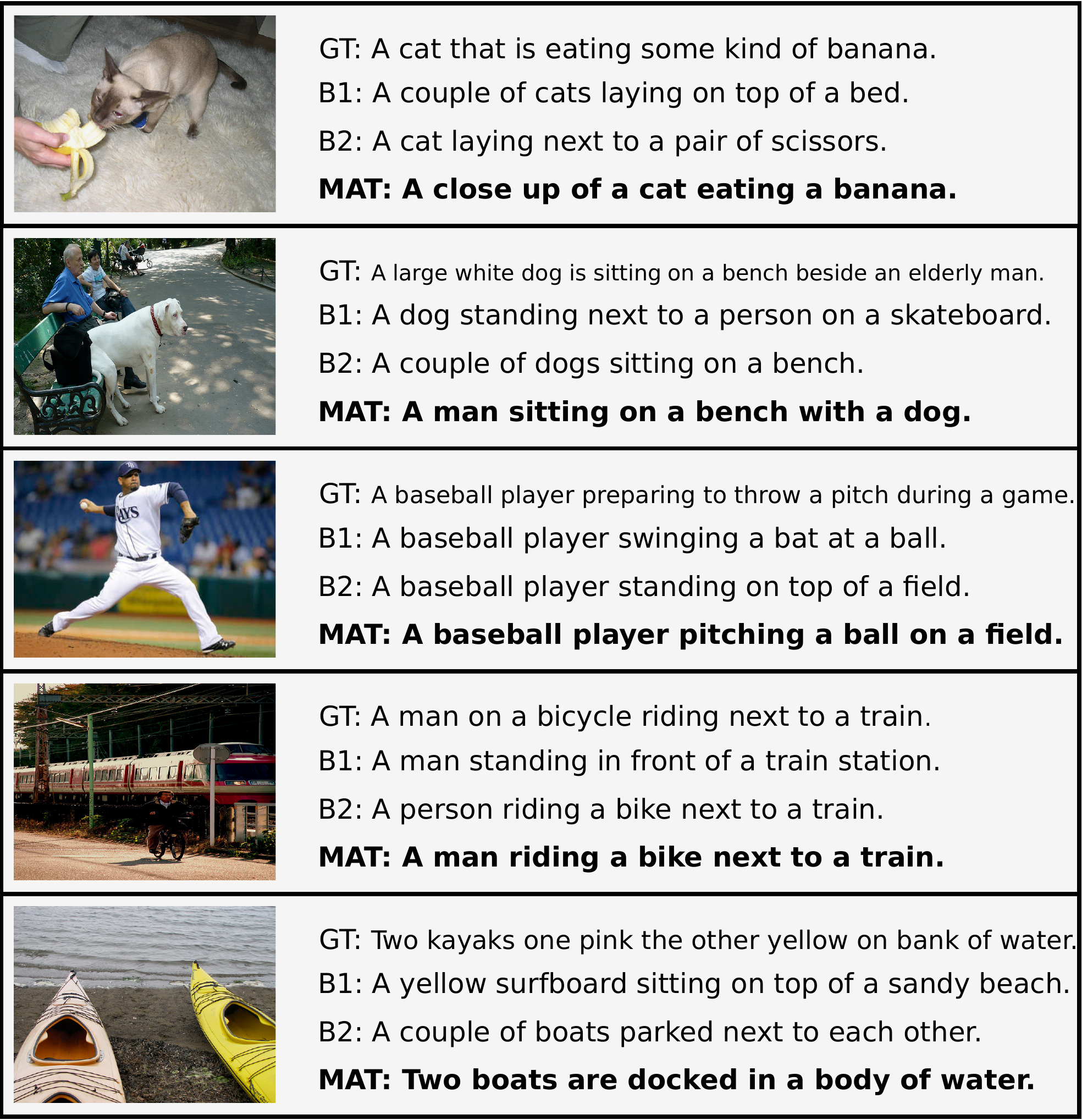}
	\end{center}
	\caption{Random selected capton generation results on MS COCO testing 5000 split. GT stands for ground truth sentence. B1, i.e., baseline1, and B2, i.e., baseline 2 shows CNN+RNN and Seq2Seq baseline methods respectively. MAT shows our final generation results.}
	\label{fig:generation_results}
\end{figure}

We present the captioning results of the proposed method in Figure\ref{fig:generation_results}. The results are randomly selected from the unused MS COCO testing 5000 split. We show four sentences for each image: GT shows the ground truth sentence; B1 shows the baseline1 method, i.e., CNN+RNN model; B2 shows the baseline2 method, i.e., Seq2Seq model (i.e. MAT w/o attention); and finally, MAT is our translation model with sequential attention layer. From the figure we can see some interesting results. Take the bottom picture for example. The baseline model 1 mistake the two boats for one boat, which MAT model predicts correctly. Moreover, the MAT model correctly predicts actions, e.g., in the middle picture, MAT predicts the action `pitching', rather than `swing' or `stand'.

\section{Conclusion}
In this work, we focus on improving the performance of image captioning by introducing two novel ideas. The first is to formulate the task as a multimodal sequence-to-sequence translation task, where the source language is represented as a sequence of objects detected from the image. The second is to introduce a sequential attention layer, which take all encoding hidden states in to consideration when generating each word. The proposed model shows superior performance over state of the art methods, and is quantitatively and qualitatively evaluated to show the effectiveness of the method.

\section*{Acknowledgments} 
This paper is jointly supported by National Natural Science Foundation of China under with Grant No.61621136008, 61327809,61210013,91420302 \& 91520201. This paper is also supported by NSF STC award CCF-1231216.

\bibliographystyle{named}

\clearpage
\bibliography{references}

\end{document}